# Improved Crowding Distance for NSGA-II

Xiangxiang Chu and Xinjie Yu
Department of Electrical Engineering, Tsinghua University, Beijing100084, China

**Abstract:** Non-dominated sorting genetic algorithm II (NSGA-II) does well in dealing with multi-objective problems. When evaluating validity of an algorithm for multi-objective problems, two kinds of indices are often considered simultaneously, i.e. the convergence to Pareto Front and the distribution characteristic. The crowding distance in the standard NSGA-II has the property that solutions within a cubic have the same crowding distance, which has no contribution to the convergence of the algorithm. Actually the closer to the Pareto Front a solution is, the higher priority it should have. In the paper, the crowding distance is redefined while keeping almost all the advantages of the original one. Moreover, the speed of converging to the Pareto Front is faster. Finally, the improvement is proved to be effective by applying it to solve nine Benchmark problems.

## I. Introduction

Multi-objective problems (MOP) are very common in our daily life, such as product design, industry producing and so on. For example, in the field of industry producing, cost and failure rate of products are both of great value. Of course, producers are earnest to see that both factors can be the minimum, which means they can get maximal profits. However, in most circumstance, they both cannot be as small as possible simultaneously. How to deal with this kind of multi-objective problem? Before NSGA-II, the weight sum method (WSM) is widely used [1]. Its basic idea is to assign a weight to each objective (factor), then minimize or maximize the sum of all objectives, which actually transforms a multi-objective problem into a single objective problem (SOP). One of the disadvantages of this method is that the weight assigned to the objective needs the expert background, which is infeasible in many circumstances. In the extreme conditions, for example, the number of objectives is more than ten; even experts cannot choose the weight ideally. Another disadvantage is that each run of WSM can provide only one solution. Therefore, it is hard to have more options.

In view of the disadvantages mentioned above, some new methods broom, among which VEGA, NSGA-II and SPEA2[1-2] are the outstanding representatives. Different from WSM, these methods deal with multiple objectives simultaneously. Parameters like weights are no more used. In recent years, these algorithms have been given more concerns and highly developed. Besides, every run of these algorithms can provide more than one acceptable solution, so they enable users to make their choices according to their preferences.

NSGA-II was suggested by professor Deb in 2002[3], which offers a new method dealing with MOP and has aroused great concern ever since. By introducing the crowding distance and the Pareto rank, NSGA-II makes sure that the final solutions have good convergence and distribution characteristic. The principle of NSGA-II will be briefly introduced later. Without loss of generality, this paper takes minimization objectives in the MOP for example. The process of NSGA-II can be divided into four steps, which is illustrated by Fig.1.

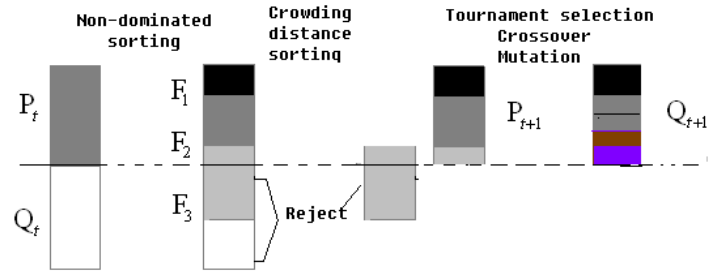

Fig.1 Procedure of NSGA-II

1) Population Initialization

There are two kinds of populations, marked with $P_t$ and $Q_t$ respectively. $P_t$ is the *t*'th generation population of $P$, and $Q_t$ is the *t*'th generation population of $Q$. Supposing that for all generation, the number of individuals in both $P_t$ and $Q_t$ is the same, marked with *N*. When *t*=1, initialize $P_1$ and $Q_1$ randomly.

2) Non-dominated sorting

The total individual number of $P_t$ and $Q_t$ is 2*N*. Sort the 2*N* individuals and assign them to different Pareto rank according to the domination relationship. $Rank_i$ is the *i*'th rank. For individuals in the different rank, the smaller its rank is, the better the individual is.

3) Selection

This step is aimed to select *N* better ones from 2*N* individuals, i.e., produce $P_{t+1}$. Selections are made according to the following criteria

First, select individuals from the smallest rank (begin with $rank_1$) to bigger rank until the number of individual is close to but not greater than *N*. Make sure the individuals on the smaller rank are all selected. Assuming that the largest selected rank is $rank_j$, the total number of individuals from $rank_1$ to $rank_j$ is *M*, therefore, it holds that $M \leq N \leq M + |rank_{j+1}|$. So the remaining *N-M* individuals are selected through the next step.

Second, this step is to select *N-M* individuals from the $rank_{j+1}$. The crowding distance is introduced to select individuals from the same rank .The larger distance one individual has, the higher priority it owns.

Take care that the crowding distance is used only when there is a must to select individuals from the same rank, i.e., the crowding distance is a criterion to select individuals within the same rank,however, rank order is the criterion to select individuals from different ranks.

4) Propagation

After $N$ individuals $P_{t+1}$ are gotten, this step is to produce other $N$ individuals. Tournament selection is widely used here which brings an outstanding advantage: parameter-less. The core of it is to make sure that $k$ individuals are selected and the chance of being selected is equal for all individuals. One common value for $k$ is 2, which is also used in this paper. Then, crossover and mutation are applied to produce $Q_{t+1}$. Any proper crossover and mutation operators can be applied here.

Among these four steps, the crowding distance plays an important role. It not only helps to select individuals within the same rank, but also makes sure the whole population are evenly distributed. Around the crowding distance, further studies are made to make NSGA-II more powerful.

## II. Crowding Distance and Improved Crowding Distance

### 2.1 Crowding Distance

From the presentation above, we cannot emphasize the importance of the crowding distance too much. For the $j$'th one in $rank_i$, its crowding distance (marked with $dis^j$) can be defined as follows(supposing that the number of objective is $m$)

1) Initialize $dis^j$ to be zero
2) For every objective function $k$
   a) Sort the individuals in $rank_i$ based on the $k$'th objective function
   b) For the $j$'th individual, if the objective function is the boundary (the minimum or the maximum), assign infinite distance to it, .i.e. $dis^j=inf$
   c) Else, suppose the sort sequence for the $j$'th individual is $n$, then
   $$dis^j = dis^j + \frac{f_{n+1}^k - f_{n-1}^k}{f_{max}^k - f_{min}^k} \quad (1)$$

The core of the crowding distance is to find the Euclidian distance between neighbor individuals in the $m$-dimension space. The individual located in the boundary is always selected because it has been assigned the infinite distance.

### 2.2 Further Consideration on Crowding Distance

From the initial definition of crowding distance above, it has explicit advantages. It ensures the better distribution characteristic from the definition itself. Moreover, the algorithm doesn't bring in any parameter, which means it is parameter-less and more universal to almost all MOPs. However, further study on the initial definition can lead to the fact that any individual in the same cubic has the same crowding distance, which has the need to be improved. In order to make a clear explanation, two objectives are taken for example.

As can be seen from Fig.2, supposing individual $E$ or $F$ might be in the dash-line cubic. According to Eq (1), they have the same crowding distance, i.e. the half circumference of the rectangle. However, obviously, the individual $E$ is better than $F$, and the former should have the priority to be chosen. However, the initial definition cannot tell apart the difference between them. That is where the idea of improvement in this paper origins from.

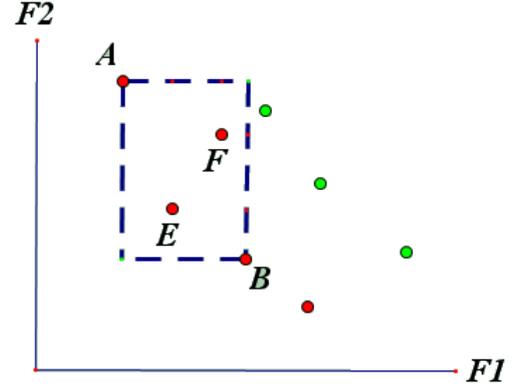

Fig.2 Demonstration about the crowding distance in the initial definition

### 2.3 Improved Crowding Distance

The discussion above discloses the direction of the improvement. After lots of attempts, a new definition of the crowding distance has been found. Just a small change will produce the new definition. The new definition for the $j$ individual in $rank_i$ can be redefined as follows:
$$dis^j = dis^j + \frac{f_{n+1}^k - f_n^k}{f_{max}^k - f_{min}^k} \quad (2)$$

Compared with the initial definition in Eq. (1), the new definition just turns the $f_{n-1}^k$ into $f_n^k$. Though it seems to be only a very small change, in fact, it has a significant effect. First, it almost inherits all advantages of the initial definition, such as parameter-less, good distribution characteristic and so on. Secondly, it adds no algorithm complexity to the initial one. Last but not least, which emphasizes the advantage of the improvement; it has the stronger capacity to converge to the Pareto Front.

In order to give a more explicit explanation, the two objective problems above are still used. As Fig.3 shows, we can easily tell the difference between the two definitions. For the initial definition, the crowding distance of $F$ is the half circumference of the outer big rectangle which is formed by $A$ and $B$, while the new crowding distance of it is the half circumference inner small rectangle. In the new definition, whether the object value lies in dot $E$ or $F$ makes difference, the former will have high priority to be selected if there is a need to choose one according to the crowding distance, which can compensates the disadvantages of the initial definition.

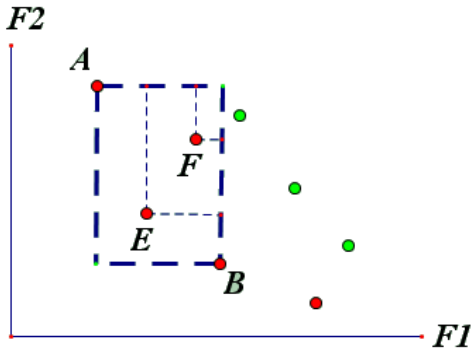

Fig.3 Illustration on the improved crowding distance

In order to demonstrate the improvement more clearly, two objective problems are used. We suppose that the objective value domain for points *A* and *B* in Fig. 2 and Fig. 3 is $[0,2] \times [0,2]$ and neglect the normalization in the denominator Eq. (1) and Eq. (2) to illustrate the key idea. The difference between the two definitions can be seen clearly from Fig.4 and Fig.5. In the initial definition, the crowding distance of any dot in the area is the same: 4, i.e. the crowding distance set of the area is a flat plane. However, the latter is different; the crowding distance set of the area is a slope. Obviously, dot (0,0) should have the highest priority with crowding distance 4, and dot(2,2) should have the lowest one with crowding distance 0.

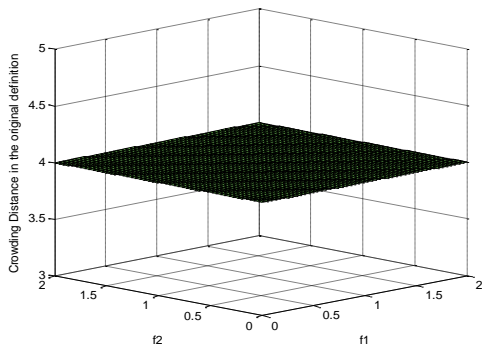

Fig.4 Crowding distance in the initial definition

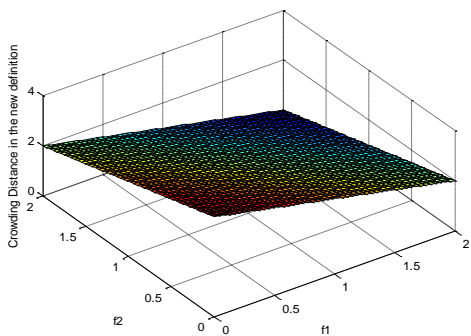

Fig.5 Crowding distance in the improved definition

A further consideration about the improved definition will come to the conclusion that the improved crowding distance gains better convergence to the Pareto Front while keeps all the advantages of the initial one.

III. Simulation results

*3.1 brief introduction*
It's essential to solve the benchmark problems in the improved definition in order to testify the validity of the rectification. Before the analysis of the simulation results, some presentations are needed.
1) Considering that the improvement focuses on the definition of the crowding distance in NSGA-II, so NSGA-II using the initial crowding distance definition is used as the control group. Moreover, except for the crowding distance, all other processes and parameters are kept the same in both groups, such as style and the rate of crossover, mutation, as well as tournament selection and so on. It can make the comparison more valid and direct. In addition, convex crossover and polynomial mutation is used. The probability of performing crossover is 0.9 and the probability of performing mutation is 0.1.
2) Considering that for multi-objective problem, to what extent the solutions converge to the Pareto Front and to what extent they are distributed should be measured simultaneously. So performance indices (PI used in the following) relevant to the two aspects should be put into use respectively. In this paper, for either aspect, two classical performances are applied to evaluate the performance.
3) Base on the same initial population, each algorithm runs 50 times. Compute the PIs in every run, then depicts the statistic characteristic so as to eliminate the effect of the random of genetic algorithm derived from the crossover and mutation process and so on. For all benchmark problems, the size of population is 50 and the final generation is 600.
4) All the 9 Benchmark problems come from [3].

*3.2 Performance Indices*
*3.2.1 Performance Indices based on convergence to Pareto Front*
In general, this kind of PIs usually focuses on the rate how many real Pareto Front solutions exist among all non-dominated solutions, or the distance how far the non-dominated solutions are from the real Pareto Front and so forth.

In 1999, Van Veldhuizen put forward an unary PI called General Distance(GD)[4],which reflects how close the non-dominated solutions is to the Pareto Front. To define GD, $d_i$ which represents the distance between one non-dominated individual *i* to the Pareto Front is defined below:

$$d_i = \min_{p \in PF} \sqrt{\sum_{k=1}^{m}(z_k^i - z_k^p)^2} \qquad (3)$$

where $z_k^i$ stands for the *k*'th objective value of the *i*'th non-dominated individual, while $z_k^p$ represents the *k*'th objective value of individual on the Pareto front. Owing to the fact that not all the problems have explicitly analytical solutions, even if exits, it's hard to find the distance, because the computation of closest distance is difficult itself. So when $d_i$ needs to be computed, usually, it can be gotten by the following steps. First, select *N* evenly-distributed solutions (an integer, usually about 1000 or larger, which depends on the requirement of precision and the limitation of time, space, etc.) on the Pareto Front. Second, get the distance between the *i*'th individual and the selected ones on the Pareto Front, and choose the shortest distance as $d_i$. From the above definition, it makes clear that $d_i$ reflects how far the *i*'th individual is away from the Pareto front. Then $d_i$ can be used to define GD, which can be written below.

$$GD(s) = \frac{(\sum_{i=1}^{|S|} d_i^q)^{\frac{1}{q}}}{|S|} \qquad (4)$$

where $|S|$ represents the number of non-dominated individual in the solution set *S*. The integer *q* determines the different distance. When *q*=2, it is the Euclidean distance, which is used in this paper. When *q*=1, then it becomes the one used in Professor Deb's paper[3].

In 1999 E.ZitZler proposed a binary PI. The binary PI called $C(S_1, S_2)$ and $C(S_2, S_1)$ is selected to strongly differentiate the initial crowding distance and the improved one[5]. $C(S_1, S_2)$ is defined as follows:

$$C(S_1, S_2) = \frac{\left| \left\{ s_2 \in S_2 \mid \exists s_1 \in S_1 : s_1 \preceq s_2 \right\} \right|}{|S_2|} \qquad (5)$$

The definition below can make it clear that what on earth the operator $\preceq$ means. Supposing $S_1$ and $S_2$ are two solution sets, *i* and *j* are individuals within these two sets respectively, the *l*'th objective value are marked with $z_l^i$ and $z_l^j$, if the following

$$z_l^i \leq z_l^j \{l = 1,2,3 \dots m\} \qquad (6)$$

is true, then we say *i* weakly dominates *j*, i.e. $i \preceq j$

From the definition, it is evident that $C(S_1, S_2)$ means the rate of the non-dominated solutions in $S_2$, which are weakly dominated by the non-dominant solutions in $S_1$. The larger $C(S_1, S_2)$ is, the better $S_1$ is. Combined with the improvement on the crowding distance, this PI can directly judge whether the improvement is successful as well as to what extend it is better than the initial definition.

*3.2.2 Performance Indices based on distribution*

Unlike single objective problem, distribution performance should be taken into account when dealing with MOP. The more evenly the solutions are distributed, the better performance they have.

In 1995, a distribution PI called spacing (SP) was put forward by J.Schott[6], which stands for the standard deviation of the closest distances. The 1-norm is used to define it.

$$d_i = \min_{s_j \in S \cap s_j \neq s_i} \sum_{k=1}^{m} |z_k^i - z_k^j| \qquad (7)$$

$$\bar{d} = \frac{\sum_{i=1}^{|S|} d_i}{|S|} \qquad (8)$$

$$SP(S) = \sqrt{\frac{1}{|S|-1} \sum_{i=1}^{|S|} (d_i - \bar{d})^2} \qquad (9)$$

where $z_k^i$ is the *k*'th objective value of the *i*'th individual, and 1-norm is used to defined $d_i$, which reflects the shortest distance between the *i*'th individual and the other individuals in *S* under the 1-norm distance. $\bar{d}$ is the average distance and the $SP(S)$ is the standard deviation of these distances. Smaller $SP(S)$ means better distribution.

As a distribution PI, $SP(S)$ puts much emphasis on the deviation of closest distance. Another PI called $M_2^*$ is also introduced for distribution, bases on the concept of niche[5].

$$M_2^* = \frac{1}{|S|-1} \sum_{i=1}^{|S|} |s_j \in S| \|s_i - s_j\| > \sigma | \qquad (10)$$

where $\sigma$ is the radius of niche, and $\| \ \|$ is the Euclidean distance. The parameter σ is sensitive, too small or too big can not get useful result. In the paper, $\sigma$ is chosen as the ten percent of the largest distance among the non-dominated individuals. Generally speaking, bigger $M_2^*$ means better distribution.

*3.3 Simulation Results*

Various Benchmark problems are chosen to verify the validity of the improvement on crowding distance used in NSGA-II. The premises are mentioned before, and NSGA-II based on the initial crowding distance acts as the control group. All are the same except the definition of the crowding distance. Two kinds, four performance indices are chosen to evaluate the validity of the improvement more comprehensively.

As is mentioned before, GD is the distance from non-dominated solutions to the Pareto Front. Smaller GD means better convergence. As Table 1 shows, the GD based on the improved crowding distance is better than the control group.

However, since the solutions are gotten by the end of 600 generation, when the control has a good convergence to Pareto Front already, so even the experiment group is better than the control group, the difference and advantage is not so evident and direct. If the generation is smaller, the difference will be more evident.

Table 1 GD index between the two definitions

| Kinds | GD (Initial CD) | GD (Improved CD) |
|---|---|---|
| SCH | 9.443e-008 | 9.369e-008 |
| FON | 0.009161 | 0.009155 |
| POL | 0.005882 | 8.446e-006 |
| KUR | 0.09277 | 0.09454 |
| ZDT1 | 0.001346 | 0.001372 |
| ZDT2 | 0.0003563 | 0.0001505 |
| ZDT3 | 0.0001007 | 4.043e-005 |
| ZDT4 | 10.90 | 10.80 |
| ZDT6 | 0.02662 | 0.0182 |

The binary performance indices can reflect the difference more directly and obviously. In Table 2, $S_1$ is the non-dominated solutions found in the initial crowding distance, while $S_2$ is gotten in the improved crowding distance. As Table 2 shows, for all Benchmarks, $C(S_2,S_1)$ is nearly twice as big as $C(S_1,S_2)$, which means the non-dominated solutions based on the improved crowding distance are better than those based on the initial crowding distance in view of convergence to the Pareto Front.

Table 2 C index between the two definitions

| Kinds | $C(S_1,S_2)$ | $C(S_2,S_1)$ |
|---|---|---|
| SCH | 0.0076 | 0.0140 |
| FON | 0.0455 | 0.1715 |
| POL | 0.0292 | 0.0896 |
| KUR | 0.1316 | 0.2476 |
| ZDT1 | 0.2012 | 0.4064 |
| ZDT2 | 0.2683 | 0.5026 |
| ZDT3 | 0.2400 | 0.6480 |
| ZDT4 | 0.2652 | 0.5808 |
| ZDT6 | 0.1272 | 0.3524 |

From the previous presentation, the improvement is valid for the enhancement of convergence. However, to evaluate the improvement more comprehensively, performance indices about distribution need to be testified. If the distribution of solutions becomes poor, then we cannot come to a conclusion that the improvement is valid. SP represents the deviation of closest distance. Smaller SP means good distribution. Table 3 shows that SP based on the improvement crowding distance is just a little larger than the control group for SCH, KUR and ZDT2. But, in fact, the gap between the initial CD and Improved CD is so small that we almost say they are the same. For other problems, the improved crowding distance behaved better. That is to say, distribution based on the improved crowding distance doesn't become worse at least.

Table 3 SP comparison between the two definitions

| Kinds | SP (Initial CD) | SP (Improved CD) |
|---|---|---|
| SCH | 0.005723 | 0.006456 |
| FON | 0.000168 | 6.117e-005 |
| POL | 0.068220 | 0.048550 |
| KUR | 0.035590 | 0.042000 |
| ZDT1 | 0.000317 | 0.000217 |
| ZDT2 | 0.000170 | 0.000231 |
| ZDT3 | 0.000251 | 0.000154 |
| ZDT4 | 0.003179 | 0.002466 |
| ZDT6 | 0.160900 | 0.112200 |

As is mention above, larger $M_2^*$ means better distribution. From Table 4, for SCH, POL KUR, and ZDT4, the $M_2^*$ using the improved crowding distance is a little smaller than the initial one, however the difference is so small that we cannot even say the initial one is better. For other problems, the improved crowding distance behaved better.

Table 4 $M_2^*$ comparison between the two definitions

| Kinds | $M_2^*$ (Initial CD) | $M_2^*$ (Improved CD) |
|---|---|---|
| SCH | 42.31 | 42.21 |
| FON | 33.65 | 33.73 |
| POL | 38.22 | 38.11 |
| KUR | 39.86 | 39.76 |
| ZDT1 | 30.04 | 32.09 |
| ZDT2 | 24.38 | 27.46 |
| ZDT3 | 40.57 | 41.07 |
| ZDT4 | 7.757 | 7.432 |
| ZDT6 | 31.55 | 34.49 |

Considering the two PIs about distribution, a conclusion can be made that NSGA-II based on the improved crowding distance is not worse than the initial one on distribution characteristic.

IV. Conclusion

In this paper, the improvement on the crowding distance in NSGA-II is demonstrated in details. The improved crowding distance overcomes the disadvantages in the initial one, i.e. the crowding distance is the same in the same cubic.

By comparing four PIs on 9 benchmark problems, we can come to the conclusion that the improved crowding distance has better convergence performance while keeping the similar distribution characteristics. It also inherit other good properties from NSGA-II, such as parameter less and universal application and so on. From the precious content, it can be seen that the improvement doesn't affect the other three steps in NSGA-II, so it is very easy to apply. There are lots of improvements on the other three steps, of which NSGA-II embed with Differential Evolution is a good example[7]. Further study will be made to testify the characteristic of combining this improvement with other improvement in the other three steps.

**Xiangxiang Chu** received the BS degree in electrical engineering from Southeast University, Nanjing, China in 2010. He is now working towards master degree in electrical engineering in Tsinghua University,Beijing,China.

**Xinjie Yu**received the BS and PhD degrees in electrical engineering from Tsinghua University, Beijing, China, in 1996 and 2001, respectively.He is an Associate Professor of the Department of Electrical Engineering at Tsinghua University. His research interests include all aspects of computational intelligence and computational electromagnetics.